\documentclass[journal]{IEEEtran}
\usepackage{times}
\usepackage{epsfig}
\usepackage{graphicx}
\usepackage{amsmath}
\usepackage{amssymb,balance }
\usepackage{bm}
\usepackage{caption}
\usepackage{amsfonts}
\usepackage{cite}

\usepackage{subfigure}
\usepackage{setspace}
\usepackage{multirow}
\usepackage{bm}
\usepackage{psfrag}
\usepackage{url}
\usepackage{stfloats}
\usepackage{tabularx}
\usepackage{array}
\usepackage{booktabs} 
\usepackage{epsfig}

\ifCLASSINFOpdf
\else
\fi
\begin{document}
%
\title{Cycle-IR: Deep Cyclic Image Retargeting}
%
%
%

\author{Weimin Tan, Bo Yan, Chumin Lin, Xuejing Niu  
\thanks{}
\thanks{}
\thanks{}}

\maketitle


\begin{abstract}
Supervised deep learning techniques have achieved great success in various fields due to getting rid of the limitation of handcrafted representations. However, most previous image retargeting algorithms still employ fixed design principles such as using gradient map or handcrafted features to compute saliency map, which inevitably restricts its generality. Deep learning techniques may help to address this issue, but the challenging problem is that we need to build a large-scale image retargeting dataset for the training of deep retargeting models. However, building such a dataset requires huge human efforts.

In this paper, we propose a novel deep cyclic image retargeting approach, called Cycle-IR, to firstly implement image retargeting with a single deep model, without relying on any explicit user annotations. Our idea is built on the reverse mapping from the retargeted images to the given input images. If the retargeted image has serious distortion or excessive loss of important visual information, the reverse mapping is unlikely to restore the input image well. We constrain this forward-reverse consistency by introducing a cyclic perception coherence loss. In addition, we propose a simple yet effective image retargeting network (IRNet) to implement the image retargeting process. Our IRNet contains a spatial and channel attention layer, which is able to discriminate visually important regions of input images effectively, especially in cluttered images. Given arbitrary sizes of input images and desired aspect ratios, our Cycle-IR can produce visually pleasing target images directly. Extensive experiments on the standard RetargetMe dataset show the superiority of our Cycle-IR. In addition, our Cycle-IR outperforms the Multiop method and obtains the best result in the user study. Code is available at \url{https://github.com/mintanwei/Cycle-IR}.
\end{abstract}


%
\IEEEpeerreviewmaketitle

\section{Introduction}

The popularity of mobile devices has greatly improved the quality, efficiency, and convenience of people life. However, the diverse display size of mobile devices leads to a media display problem that due to the different size between the input media and the device screen, media may not be perfectly suitable for full-screen display. With the explosive growth of media content on social networks, this problem is further aggravated. Traditional image resizing methods often cause shrinking, stretching, or clipping. To solve this problem, content-aware image retargeting approaches are proposed to manipulate media content to make it adapt to different aspect ratios of device screen intelligently\cite{Liang2017ObjectiveQP,Dong2016ImageRB,Zhou2017PerceptuallyAI,Shao2017QoEGuidedWF,Lau2018ImageRV,Li2019LearningAC,Lin2014ObjectCoherenceWF,Lei2017DepthPreservingSI,Li2015DepthPreservingWF}. Recently, image retargeting algorithms are also extended to various fields such as aesthetic enhancement\cite{Xiang2010VideoRF}, video synopsis\cite{Li2016SurveillanceVS,Nie2013CompactVS}, and low bit-rate image retrieval\cite{Tan2016ImageRF,Tan2018BeyondVR,Yan2017CodebookGF}, \textit{etc}.

\begin{figure}[t]
\centering
\includegraphics[width=8.5 cm]{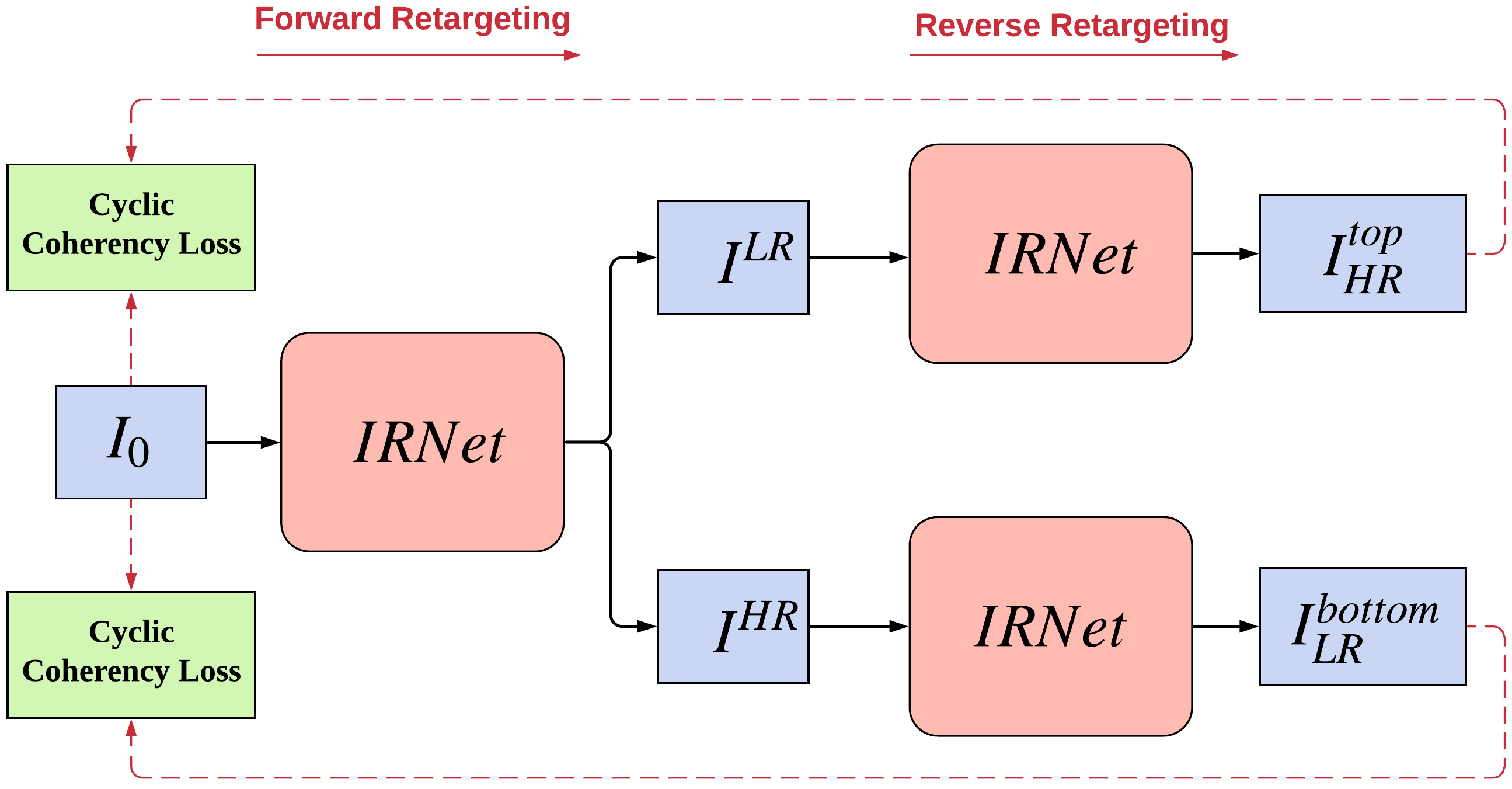}
\caption{Overview of the proposed Cycle-IR framework. By cyclically utilizing the retargeting results of the forward inference of deep image retargeting network (IRNet), our IRNet can be trained in an unsupervised way, without the requirement of any manual label.}
\label{FigOverFramework}
\end{figure}

A good content-aware image retargeting approach is able to preserve visually important object details, while removing unimportant pixels and preventing obvious distortions. Numerous works have been proposed and achieved great success in recent years\cite{avidan2007seam, fang2011saliency, mansfield2010scene, qi2012seam, noh2012seam, dong2014summarization, grundmann_2010, panozzo2012robust, zhang2015retargeting}. Typically, these image retargeting methods follow a conventional pipeline. Firstly, they calculate a saliency map to measure the importance of each pixel in the input image. Then, several constraints such as content and structure similarity or symmetry preserving are defined. Finally, based on the calculated saliency map and the defined preserving constraints, they define a manipulating way to remove or insert pixels in the input image, yielding the desired target image. Such a strategy is limited by the fixed design principle. For example, in traditional retargeting algorithms, saliency map is widely employed and calculated based on the low-level features of input image such as image edge, color contrast, depth information, or handcrafted features. The failure of saliency map estimation commonly results in poor quality of retargeting results, which greatly limits the generality of these approaches.

Deep learning techniques have demonstrated outstanding performance in various fields. However, few studies have attempted to solve the image retargeting problem using deep learning techniques. The reason is that image retargeting task belongs to an uncertain problem, \textit{i.e.}, the way to produce the ideal retargeting results is uncertain. In addition, the best way to evaluate retargeting results is subjective evaluation. These reasons make the image retargeting task quite different from those tasks with a definite goal or objective evaluation criteria, such as image super-resolution, image classification, object detection, and semantic segmentation, \textit{etc}. Furthermore, these reasons also make it difficult to build a large-scale image retargeting dataset that contains input images and corresponding target images with different aspect ratios.

Without such a dataset, we cannot accurately train an image retargeting model that is based on conventional deep learning techniques. Therefore, using deep learning techniques to solve image retargeting problem becomes an extremely challenging problem. Cho $et~al$. \cite{cho2017weakly} make an attempt to use deep learning technique to addressing image retargeting problem. They propose a weakly and self-supervised deep network to learn a shift map for manipulating the aspect ratio of input images. In order to obtain a good shift map, they use two loss terms of structural similarity and content consistency to optimize the deep network. Despite the encouraging progress, this method requires the input image to be of fixed size, which greatly limits the practicability of this method.

In this study, we solve the aforementioned problems by introducing a deep cyclic image retargeting approach (\textbf{Cycle-IR}). Our key idea is that whether the size of the input image is reduced or increased, the target image generated by an ideal image-retargeting model should be highly consistent with the visual perception of the input image, free of distortions and artifacts, and complete preserving of important visual information, \textit{etc}. Therefore, if the retargeted image is fed back to the ideal image-retargeting model as input, the mapped-back image should be similar to the original input image in terms of visual perception and structural similarity. The coming questions are \textit{\textbf{how to establish the forward-backward mapping consistency?}} and \textit{\textbf{how to effectively evaluate the visual consistency between the input image and the mapped-back image?}}

In light of the above discussions, this paper introduces a novel Cycle-IR approach based on the characteristics of image retargeting that is able to generate target images with different resolutions. Accordingly, we propose a deep image retargeting network (\textbf{IRNet}) to implement the retargeting process. In order to form the cycle relation, our IRNet is designed to output a low-resolution and a high-resolution target image, simultaneously, which is different from conventional retargeting methods that output only a single target image. Fig. \ref{FigOverFramework} shows the basic idea of our Cycle-IR, which consists of a two-stage inference process. The first stage is to produce two target images with different resolutions given an input image and the target aspect ratio. Then, at the second stage, the image retargeting model (sharing network parameters at the first stage) uses the retargeted image as input to reconstruct the original input image. Specifically, the IRNet maps the input images to the retargeted images, and then maps them back. In order to measure the forward-backward mapping coherency effectively, we propose a cyclic perception coherence loss. This cycle coherence loss can encourage the IRNet to discover the visual importance of the input image and reduce the distortions and artifacts of the target image. In addition, this loss prevents the training process from requiring any explicit user annotations. In the entire training process of IRNet, we only need unlabeled color images and the pre-trained VGG16 model\cite{Simonyan2015VeryDC}.

We present extensive experiments to evaluate the effectiveness of our Cycle-IR on the public RetargeMe dataset. Numerous representative image retargeting methods are compared with the proposed approach explicitly in terms of visual quality and user study. To compare with the weakly supervised deep retargeting method\cite{cho2017weakly}, we implement it using Tensorflow platform. Note that our Cycle-IR is an unsupervised deep retargeting approach and is trained on the RGB images from HKU-IS dataset\cite{li2016deep} that have no overlap with RetargeMe. Therefore, all images in RetargeMe can be used as test set. Experiment results show that our Cycle-IR can implicitly learn a good mapping from input images to target images, where the mapping is able to make the input and output follow the same distribution. Besides, we conduct extensive experiments to analyze the performance of our Cycle-IR. Benefiting from the automated learning ability of deep networks, our Cycle-IR produces the target image with better visual quality than other retargeting methods, and obtains the best result in the objective evaluation of user study.

In summary, the main contributions of this work are:

\begin{itemize}
\item This work proposes a novel unsupervised deep image retargeting approach by leveraging cyclic visual perception consistency between the input images and the reverse-mapped results. This is the first attempt to address the problem of image retargeting in an unsupervised deep learning way. Our Cycle-IR does not require any labeled data, additional parameter settings, or human assistance. It performs favorably against the state-of-the-art image retargeting approaches.
\item Different from previous retargeting methods that output one retargeted image, our deep image retargeting model outputs a pair of retargeted images. This special design is helpful for obtaining a stable retargeting model by exploiting a pair of cycle constraints.
\item A cyclic perception coherence loss is proposed to evaluate the cycle coherence between the forward-backward mapping results. This cycle loss allows our IRNet to train in an end-to-end manner and enforces IRNet to implicitly learn a mapping that the input images and the target images follow the same distribution. Besides, the advantage of our cycle loss makes it possible to incorporate it into other image retargeting methods, improving the perceptual quality of the target images they generate.
\item We design a spatial and channel attention layer, which is able to discover the visually interesting areas of input images effectively, especially in cluttered images. We exploit this attention layer as a learnable network component to help assist the IRNet in learning accurate attention map.
\end{itemize}

\section{Related work}

In this section, we review previous works related to this paper. Over the past decade, numerous image retargeting works have been proposed \cite{avidan2007seam,fang2011saliency,mansfield2010scene,qi2012seam,noh2012seam,dong2014summarization, grundmann_2010, panozzo2012robust,zhang2015retargeting,Zhang2015RetargetingSP,Qu2013ContextAwareVR,Gallea2014PhysicalMF}. The majority of image retargeting methods are based on hand-crafted visual attention map such as gradient map, color contrast, object segmentation, face detection, \textit{etc}. We refer the readers to survey literature \cite{Pal2016ContentAwareIR} for more details. Here, we mainly discuss representative image retargeting methods and a recent deep learning based method. Besides, we will review the successful applications of cycle constraint in other fields.

\textbf{Conventional retargeting}: Most of conventional image retargeting approaches can directly retarget input images to target aspect ratios according to the previous calculated visual attention map, in which the bright areas indicate the interesting areas in the input image. These approaches commonly exploit traditional methods to extract pixel saliency information, and then define a manipulating way to insert or remove pixels in order to obtain the target image. Seam carving \cite{avidan2007seam} is a good example, which uses three ways ($e.g.$, pixel gradient, entropy energy, visual saliency) to measure the importance of pixels in the input image. Then, it defines a way of seam carving to change the aspect ratio of the input image. A seam is an 8-connected path of pixels from each column or row in the source image. The retargeted image is obtained by iteratively removing or inserting seams.

Different from the seam carving method in a discrete way, Panozzo $et~al.$ \cite{panozzo2012robust} propose to continuously transform the input image into the target image. Specifically, they overlay a uniform grid on the input image and assign different scaling factors to each grid cell. The scaling factors of important regions are large, and conversely, small. As a result, those important regions are preserved as much as possible, while less important regions are shrinking obviously. Wang $et~al.$ \cite{wang2008optimized} propose a scale-and-stretch warping method. They calculate optimal scaling factors for image regions based on the edge and saliency map. Jin $et~al.$ \cite{jin2010nonhomogeneous} cast image retargeting problem into a quadratic program based on a triangular mesh. All these conventional approaches need to calculate an importance map based on low-level visual cues or other human priors. The poor quality of importance map significantly degrades the performance of these methods, which seriously limits their generality.

\textbf{Weakly supervised retargeting}: Because the visual quality of retargeted images highly depends on the subjective assessment, it is hard to build a large-scale image retargeting dataset for supervisory learning a deep model. Thus, there are few supervised retargeting methods. Cho $et~al$. \cite{cho2017weakly} introduce deep learning technique into the image retargeting task. Their deep model learns a shift map to implement pixel-wise mapping from the source image to the target image. They define a loss function containing two terms: content loss and structure loss. The content loss enables the retargeted image outputted by the deep network to have the same class as the source image. Therefore, their approach requires image-level annotations to construct the content loss. This approach may fail to deal with those images that do not belong to any class in the training set. Besides, in order to reduce distortions in the target image, the method proposes to use 1D duplicate convolution to smooth the learned shift map. This operation is helpful for reducing visual artifacts in the target image but requires the input image to be a fixed size.

\textbf{Cycle Constraint}: Cycle constraints have been explored in other fields in order to regularize model predictions and improve its stability. For language translation, Richard shows the translation quality can be improved effectively by using a back translation and reconciliation strategy\cite{Brislin1970}. For computer vision, high-order and multimodal cycle consistency have been used to improve the stability and quality of model predictions such as co-segmentation\cite{Huang2013ConsistentSM}, motion prediction\cite{Zach2010DisambiguatingVR}, 3D shape matching\cite{Huang2013ConsistentSM}, semantic alignment\cite{Zhou2015FlowWebJI}, depth estimation\cite{Godard2017UnsupervisedMD}, image-to-image translation, video frame interpolation\cite{Liu2019DeepVF}. For deep learning, researchers have incorporated the concept of cycle consistency into the regularization of optimizing deep networks\cite{Zhou2016LearningDC}. In this work, we demonstrate a novel and feasible way of exploiting the concept of circle consistency to address the image-retargeting problem. To the best of our knowledge, we are the first to improve image retargeting by using circle consistency. Our Cycle-IR is capable of producing high-quality target images and achieves the state-of-the-art result.

\section{Deep Cyclic Image Retargeting}

Our goal is to develop a deep image-retargeting model that can directly generate high-quality target images when users provide input images with arbitrary sizes and desired aspect ratios. In this section, firstly, we present an overview of the Cycle-IR framework in Section \ref{OverviewCyclicIR}. Then, we introduce the cyclic perception coherence loss in Section \ref{CyclicPerceptionCoherence}. Furthermore, we introduce implementation details of the proposed IRNet in Section \ref{IRNetImplementation}. Finally, we justify our design strategy and describe the training process in Section \ref{Justifications} and \ref{TrainingandInference}, respectively.

\subsection{Cycle-IR Framework}  \label{OverviewCyclicIR}

An ideal image retargeting model should have the ability to preserve visually interesting areas in the input image and prevent significant distortions and artifacts in the target image when changing the aspect ratio of the input image. That is to say, the removed or inserted information in the target image is trivial and can be easily restored or eliminated. Inspired by this observation, we propose a cycle image-retargeting framework. The key idea of our Cycle-IR is that when the target image is obtained by eliminating the content of input image, the removed information can be effectively restored by re-feeding the target image into the retargeting model. Similarly, when the target image is obtained by inserting the content of the input image, the inserted information can be easily removed by re-feeding the target image into the retargeting model. Both cases are required to satisfy the condition that the reverse mapping result of the target image has the same pixel distribution as the input image, \textit{i.e.}, cyclic perception coherence. This cycle coherence is also helpful for obtaining target images that are free of distortions and artifacts.

\begin{figure*}[t]
\centering
\includegraphics[width=18 cm]{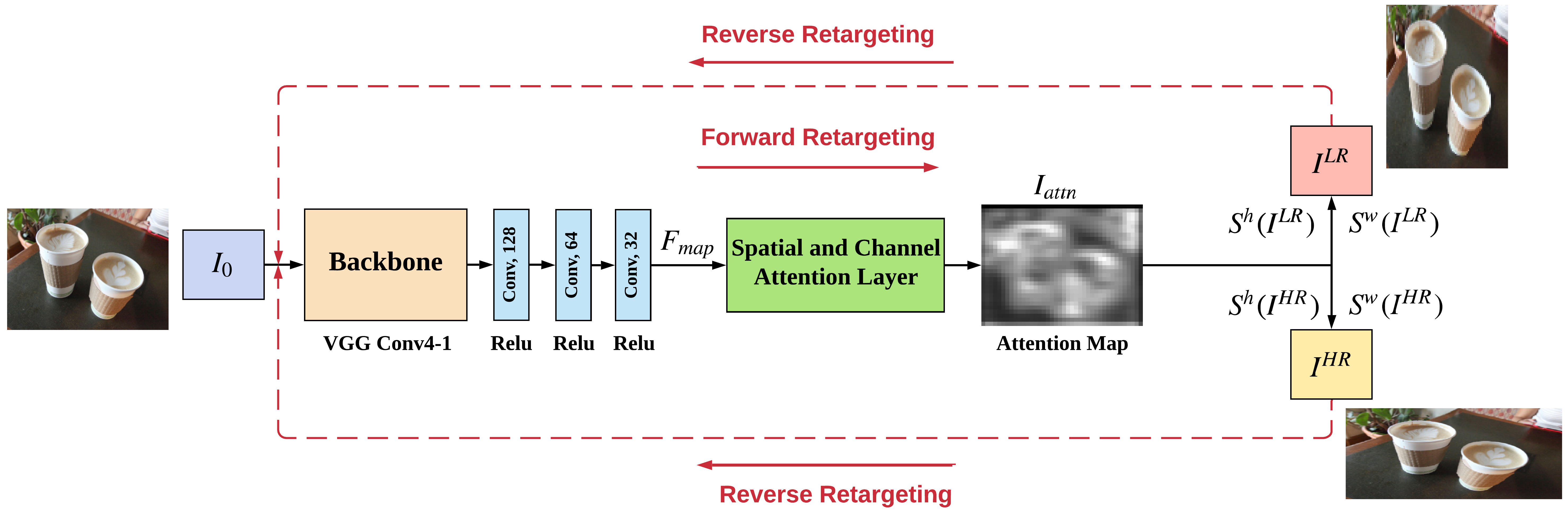}
\caption{Demonstration of the proposed IRNet architecture for implementing our Cycle-IR. The IRNet consists of a backbone (Conv4-1 of VGG-16), three convolutional layers, and a spatial and channel attention layer.}
\label{FigNetworkArchitecture}
\end{figure*}

The proposed cyclic image retargeting framework is illustrated in Fig. \ref{FigOverFramework}. We design a two-stage inference procedure so that the retargeting model shared by both forward and reverse mapping directions in the cycle constraint can be learned stably. Consider an input image $I_0$ ($H_{I_0}\times W_{I_0}$) and desired aspect ratio $\phi_h \in (0,1]$ and  $\phi_w \in (0,1]$ that are passed through the IRNet model (see Section \ref{IRNetImplementation} for detail). At the first stage (forward retargeting), the IRNet model outputs two target images $I^{LR}$ and $I^{HR}$ in order to establish the cycle relation between the forward and backward retargeting procedure. Note that this is different from previous image retargeting methods that yield only one target image in a single inference.

\begin{equation}
  I^{LR}, I^{HR} = IRNet_{FWD}(I_0)
 \label{equationFWD}
\end{equation}
The resolution of $I^{LR}$ is defined as follows:

\begin{align}
H_{I^{LR}} = \phi_h * H_{I_0} \\ \nonumber
W_{I^{LR}} = \phi_w * W_{I_0}
  \label{equationSizeOfLR}
\end{align}

Different from conventional retargeting methods, we set the resolution of $I^{HR}$ as follows:
\begin{align}
H_{I^{HR}} = \frac{\mu_h}{\phi_h} * H_{I_0} \\ \nonumber
W_{I^{HR}} = \frac{\mu_w}{\phi_w} * W_{I_0}
  \label{equationSizeOfHR}
\end{align}
where $\mu_h$ and $\mu_w$ are used to control the size of $I^{HR}$ and are set to 1 in the implementation. We design such resolutions for $I^{LR}$ and  $I^{HR}$ in order to form the circle constraint for optimizing the IRNet.

At the second stage (reverse retargeting), we re-feed the target image obtained in the first stage into the IRNet to obtain the reverse mapping results. Note that the IRNet in this stage shares the same network parameters as the first stage. The IRNet outputs $I^{top}_{LR}$ and $I^{top}_{HR}$ from the input of $I^{LR}$ and $I^{bottom}_{LR}$ and $I^{bottom}_{HR}$ from the input of $I^{HR}$, respectively.

\begin{align}
 I^{top}_{LR}, I^{top}_{HR} = IRNet_{REV}(I^{LR})  \\ \nonumber
 I^{bottom}_{LR}, I^{bottom}_{HR}  = IRNet_{REV}(I^{HR})
\end{align}

where $I^{top}_{HR}$ ($H_{I_0}\times W_{I_0}$) and $I^{bottom}_{LR}$ ($H_{I_0}\times W_{I_0}$) are used to compute the cyclic coherency loss. Note that $I^{top}_{LR}$ ($H_{I_0}*\phi_h^2\times W_{I_0}*\phi_w^2$) and $I^{bottom}_{HR}$ ($H_{I_0}*\mu_h^2/\phi_h^2\times W_{I_0}*\mu_w^2/\phi_w^2$) are not used to compute the cycle loss because their resolutions are different from the size of input image.

Finally, we can establish the cyclic perception coherence for $I_0$ and $I^{top}_{HR}$, and $I_0$ and $I^{bottom}_{LR}$. The following section will dedicate to state the cyclic perception coherence loss.

\subsection{Cyclic Perception Coherence Loss}
\label{CyclicPerceptionCoherence}

These widely used loss functions, such as the mean absolute error (MAE) and mean squared error (MSE), are not fully consistent with human visual perception when evaluating the quality of retargeted images. Minimizing the MSE loss often results in over-smoothed results because among all possible results, generative models will produce an average result to satisfy the minimum value of MSE. Besides, the pixel-level MSE considers that all pixels in the input image have the same visual importance. It will force the IRNet model to learn a uniform scaling, rather than the non-uniform resizing we desire. To solve this problem, we present the cyclic perception coherence loss $\mathcal{L}_{pair}$ to assess the perceptual consistency between the input image and the reverse-mapped image. This cycle loss enables our IRNet to implicitly discover the visually interesting areas in the input image and effectively learn a mapping that the input images and the target images follow the same distribution. In this way, we can obtain favorable retargeting results by using $\mathcal{L}_{pair}$ to optimize our IRNet.

\begin{figure*}[t]
\centering
\includegraphics[width=15.5 cm]{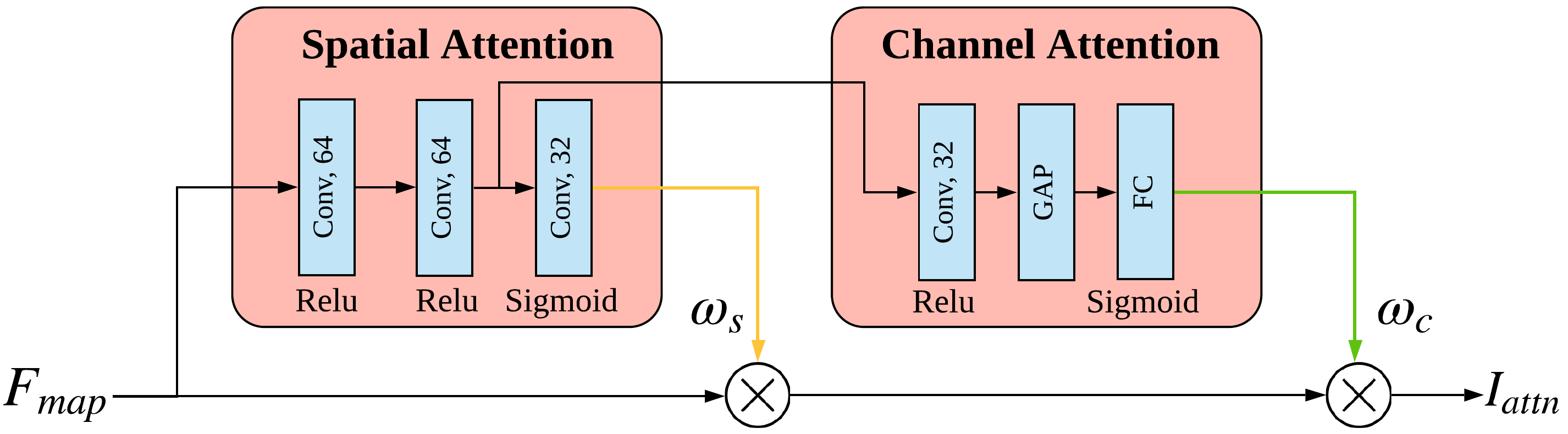}
\caption{Illustration of the spatial and channel attention layer $\Gamma_{attn}$ (corresponding to the green block in Fig. \ref{FigNetworkArchitecture}) that consists of two components of spatial attention and channel attention. The deep representation $F_{map}$ is passed through three convolutional layers to obtain the spatial weight $\omega_s$. Then, the output of the second convolutional layer in the spatial attention component feeds into the channel attention component, producing the channel weight $\omega_c$. Finally, the visual attention map $I_{attn}$ is obtained by multiplying $F_{map}$ with $\omega_s$ and $\omega_c$, see Eq. (\ref{SCALayer}) for a formal definition.}
\label{SpatialChannelAttention}
\end{figure*}

The perceptual loss has demonstrated superior performance in various fields. Different from previous approaches, we propose a pair of cyclic perception coherence loss $\mathcal{L}_{pair}$ to provide a more stable constraint for optimizing IRNet. $\mathcal{L}_{pair}$ is defined as.

\begin{align}
 \label{LossFuntion}
 \mathcal{L}_{pair}=\frac{ 1}{L }  \sum_{l=4}^L[(f_l(I_0)-f_l(I_{HR}^{top})) \times\beta_l]^2 + \\ \nonumber  \frac{ 1}{L }  \sum_{l=4}^L[(f_l(I_0)-f_l(I_{LR}^{bottom})) \times\beta_l]^2
\end{align}
where $f$ represents the pre-trained VGG16 model and $L=5$ (corresponding to the layers of VGG16 model). Considering image retargeting needs to focus not only on the semantic region of objects, but also on the overall structure of the image, we use multiple deep representations to evaluate the perceptual consistency instead of a single deep representation. $\beta_4$ and $\beta_5$ are set to 1 and 3, respectively. We empirically set $\beta_5$ to 3 because representations in deeper layer contain more semantic information. Eq. (\ref{LossFuntion}) implies that these high-level deep representations of the input image and the reverse-mapped image should be consistent.

\subsection{Implementation of IRNet}
\label{IRNetImplementation}

The proposed Cycle-IR framework in Fig. \ref{FigOverFramework} is flexible, allowing users to freely design the required network structure to implement it. Fig. \ref{FigNetworkArchitecture} shows the implementation of our image retargeting network (IRNet), which is a fully convolutional architecture. The IRNet consists of a backbone (Conv4-1 of VGG16), three convolutional layers, and a spatial and channel attention layer. Despite the simplicity of our IRNet, it achieves excellent performance in manipulating the aspect ratio of input images. More advanced networks can be easily incorporated to achieve better performance.

\textbf{Obtaining Visual Attention Map}. Given an input image $I_0$ and a target aspect ratio $\phi_h$ and  $\phi_w$, a deep representation $F_{map}$ is obtained by passing through the backbone and three convolutional layers that implement a non-linear transformation. Since the activation values of neurons in $F_{map}$ are often scattered in space and channel, we need an effective way to obtain the visual attention map corresponding to the input image. Therefore, we feed the $F_{map}$ into the spatial and channel attention layer to obtain the visual attention map $I_{attn}$.

\begin{equation}
I_{attn}=\Gamma_{attn}(F_{map})
 \label{VisualAttentionMap}
\end{equation}
where $\Gamma_{attn}$ denotes the non-linear transformation of the spatial and channel attention layer. See below for details.

\textbf{Spatial and Channel Attention Layer.} Fig. \ref{SpatialChannelAttention} shows the network architecture of the proposed spatial and channel attention layer, which consists of two components of spatial attention and channel attention. The input of channel attention component is the output of the second convolutional layer in the spatial attention component instead of the extracted deep representation $F_{map}$. This architecture design helps to utilize differentiated information from spatial attention components. $I_{attn}$ can also be calculated as follows:

\begin{equation}
I_{attn}= F_{map} * \omega_s * \omega_c
 \label{SCALayer}
\end{equation}
where $\omega_s$ and $\omega_c$ represent the spatial and channel weights produced by the spatial and channel attention components, respectively.

\textbf{Generating desired target image}. To obtain the desired target image, we employ a continuous method to deform the resolutions of input images with the guidance of the obtained visual attention map $I_{attn}$. Note that the entire deformation process is integrated into the optimization of IRNet model, which is conductive to allowing the IRNet to decide on its own way to generate the best target image. Specifically, we cover the input image with a uniform grid of M-column and N-row. The size of each grid cell is $H_{I_0}$/M $\times$ $W_{I_0}$/N. Based on the learned attention map $I_{attn}$, we calculate the scaling factor of each grid cell for reconstructing $I^{LR}$ as follows.

\begin{align}
S^h_{i}(I^{LR})= \frac{1}{N}\sum_{j=1}^{N} \frac{1}{1+e^{-I_{attn}(i,j)}}  \\ \nonumber
S^w_{j}(I^{LR})= \frac{1}{M}\sum_{i=1}^{M} \frac{1}{1+e^{-I_{attn}(i,j)}}
 \label{DesiredTargetImage}
\end{align}

Based on the calculated $S^h_{i}(I^{LR}) \in (0,1]$ and $S^w_{j}(I^{LR}) \in (0,1]$, we can easily obtain the scaling factor of each grid cell for reconstructing $I^{HR}$ as follows.

\begin{align}
S^h_{i}(I^{HR})= 1-S^h_{i}(I^{LR})+\psi_h  \\ \nonumber
S^w_{j}(I^{HR})= 1-S^w_{j}(I^{LR})+\psi_w
 \label{DesiredTargetImage2}
\end{align}
where $\psi_h$ and $\psi_w$ are the adjustment factors and set to 1 in the implementation. Generally, deep network has the ability to learn suitable adjustment factors (\textit{i.e.}, $S^h(I^{LR})$ and $S^w(I^{LR}))$ for different scenarios.

Afterwards, we deform each grid based on the calculated $S(I^{LR})=\{S^h(I^{LR}),S^w(I^{LR})\}$ and $S(I^{HR})=\{S^h(I^{HR}),S^w(I^{HR})\}$ and obtain the target images $I^{LR}$ and $I^{HR}$. It turns out that our Cycle-IR can greatly improve the quality of image retargeting without incurring extra human assistance or weak tag. Furthermore, the concept of cycle coherency is extended by considering task-specific knowledge, and can automatically decide the areas to be deleted or reserved in the input image.

\begin{figure*}[t]
\centering
\includegraphics[width=18 cm]{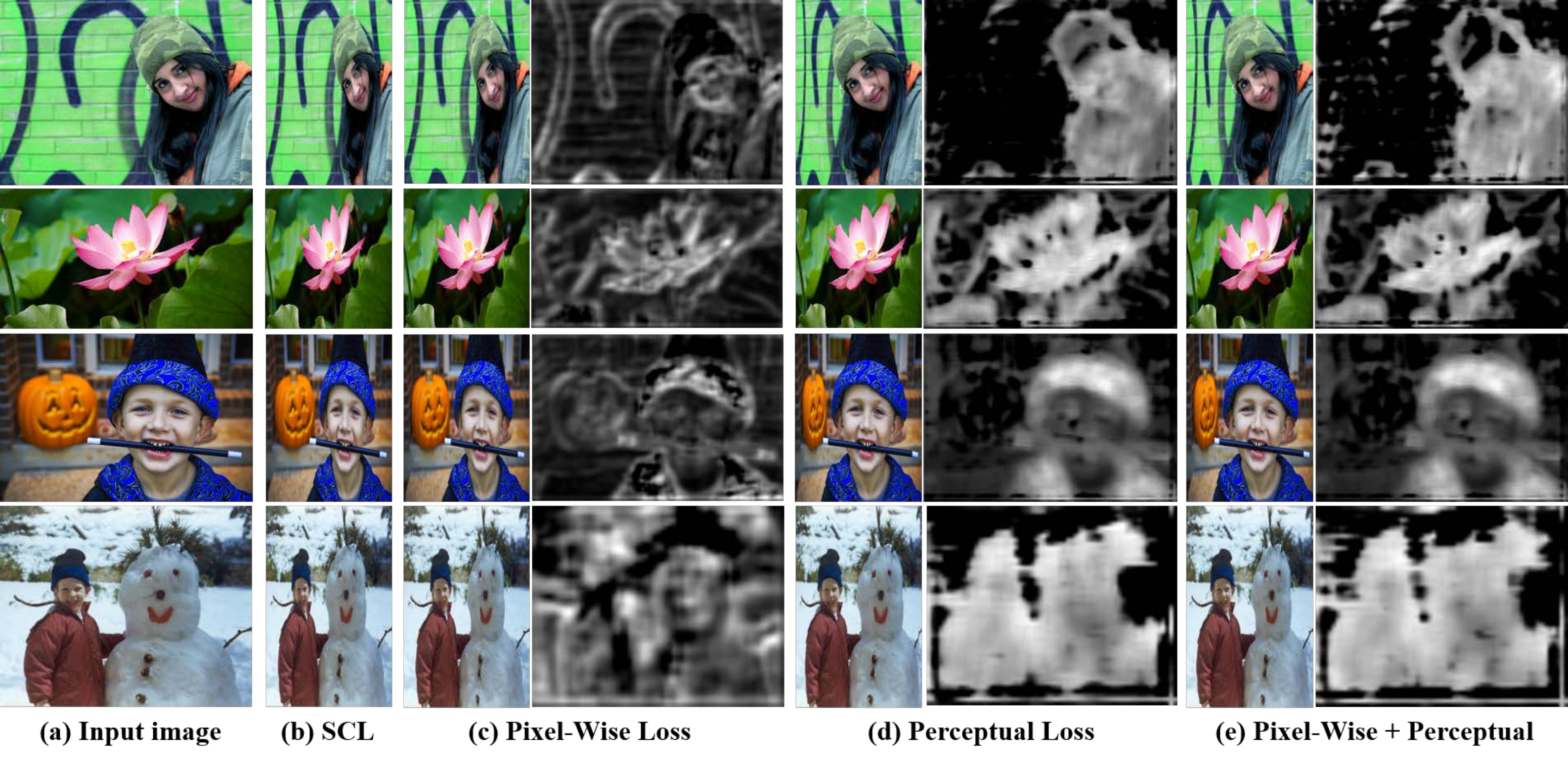}
\caption{Visual comparison of pixel-wise loss, perceptual loss, and pixel-wise plus perceptual loss to understand their impact on the optimization of Circle-IR.}
\label{FigPixelVsPerceptual}
\end{figure*}

\subsection{Justifications}
\label{Justifications}

The proposed Cycle-IR framework is motivated by the following ideas:

\begin{itemize}
\item Building a large-scale image retargeting dataset for training a deep retargeting network is a challenging work. It not only requires finding an effective way to generate target images, but also needs huge human efforts to assess their quality. In addition, to improve the practicability of image retargeting, it is necessary to get rid of the limitation of handcrafted design principles. Image retargeting algorithm needs to obtain the generalization ability of dealing with different complex scenarios by learning. Hence, we avoid the supervised and weakly supervised methods and present an unsupervised way to learn a deep retargeting model effectively.
\item Using cycle constraint to optimize the retargeting model makes deep learning based image retargeting possible. Furthermore, through cyclic perception coherence, the IRNet model can stably learn a favorable retargeting mapping, yielding visually pleasing retargeting results.
\item Visual attention areas in the input image should be located accurately, especially in cluttered images. To this end, we design a spatial and channel attention layer to find visually important areas in a learnable way. We leverage this attention layer to help assist the IRNet in learning accurate attention map. Developing such attention layer has the additional advantages of avoiding using large-scale networks to achieve similar performance, thereby reducing computational cost and memory footprint.
\end{itemize}

\subsection{Training and Inference}
\label{TrainingandInference}

To train the IRNet model, we use the RGB images from HKU-IS dataset \cite{li2016deep} as the training set. This dataset contains 4,447 RGB images. Note that saliency labels in the HKU-IS dataset \cite{li2016deep} are discarded during training, \textit{i.e.}, any RGB image without user annotations can be exploited as our training set. The RGB images from HKU-IS dataset \cite{li2016deep} contains diverse scenes, which helps to make the trained model have good generality. During training, the input aspect ratios are randomly generated for each training batch within $(H_{I_0}/4 \sim H_{I_0}/2) \times (W_{I_0}/4 \sim W_{I_0}/2)$, similar to \cite{cho2017weakly}. The training process is fast, which takes around 1 hour for 50 epochs on a machine with an Nvidia GPU Quadro M4000. Generally, after training 50 epochs, the IRNet is capable of producing satisfactory results. The loss function of Eq. (\ref{LossFuntion}) is optimized using Adam algorithm with an initial learning rate of $1\times e^{-3}$. The batch size is set to 4. During inference, given an input image and a desired target size, the IRNet model can directly output the target image with the desired aspect ratio, so no pre-processing (\textit{e.g.}, computing saliency map, detecting faces, semantic segmentation, \textit{etc.}) or post-processing (\textit{e.g.}, stitching, zooming, etc.) is necessary.

\begin{figure*}[t]
\centering
\includegraphics[width=18 cm]{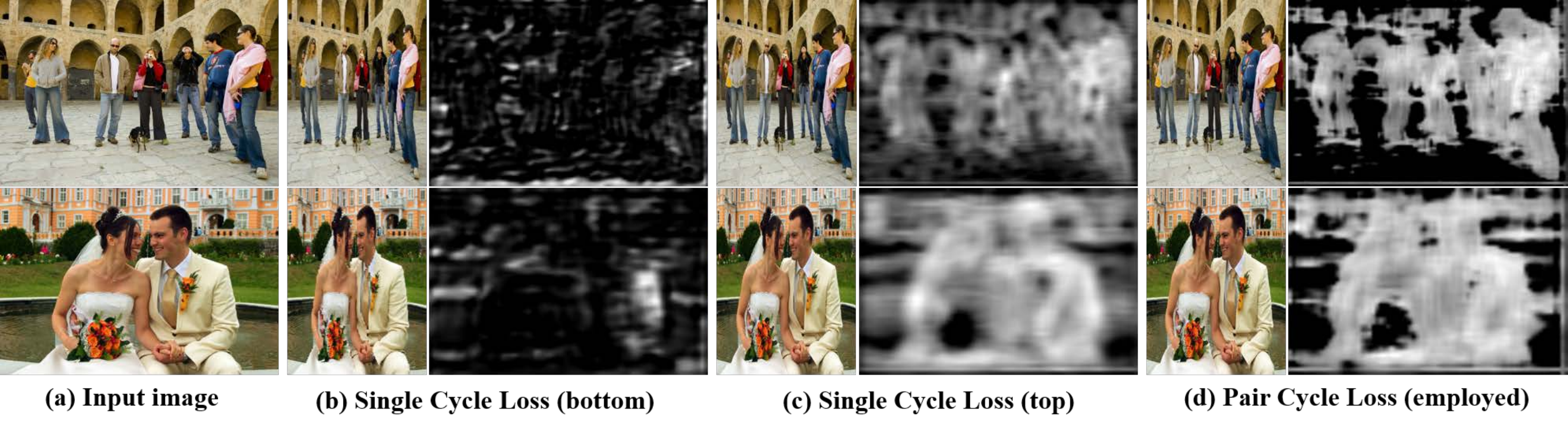}
\caption{Visual comparison of single cycle loss vs. pair cycle loss. Intuitively, ``Single Cycle Loss (top)'' poses stronger constraints than ``Single Cycle Loss (bottom)'', because restoring the information (reverse mapping in the top) removed in the forward retargeting is more difficult than removing  the information (reverse mapping in the bottom) inserted in the forward retargeting.}
\label{FigdoubleCycleLoss}
\end{figure*}

\section{Experiment}

For a clear understanding of the proposed approach, we conduct extensive evaluations to investigate the performance of the proposed Cycle-IR in Sections \ref{PixelVsPerceptual} to \ref{NetworkAndEfficiency}. Specifically, we analyze the retargeting quality of Cycle-IR under different loss functions such as pixel-wise loss vs. perceptual loss (see Section \ref{PixelVsPerceptual}) and single cycle loss vs. pair cycle loss (see Section \ref{SingleCycleVsDoubleCycle}), which help to understand the impact of different losses. In addition, we discuss the effect of human annotated saliency map on the training IRNet model in Section \ref{SaliencyGuidance}. This discussion will help us understand the role of incorporating object location information to help the IRNet model distinguish the visually important areas in the input image. Furthermore, we demonstrate several visual examples to evaluate the retargeting quality of IRNet model under different input aspect ratios in Section \ref{AdjustmentOfAspectRatio}. Finally, we introduce the network size and computational efficiency of IRNet in Section \ref{NetworkAndEfficiency}. Afterward, in Section \ref{ComparisonWithPriorArt}, we explicitly compare our model with current state-of-the-art methods on the standard RetargetMe \cite{rubinstein2010comparative} dataset in terms of visual quality and user study. Finally, in Section \ref{FailureCaseAndFutureWork}, we show several failure cases of our IRNet and provide some insights to improve it in future work.

\subsection{Pixel-Wise Loss vs. Perceptual Loss}
\label{PixelVsPerceptual}

To exhibit the difference between the pixel-wise loss and the perceptual loss, where both of them can be used to measure the difference between the input images and the reverse-mapped images, we select multiple representative images from RetargetMe dataset and show the visual comparisons in Fig. \ref{FigPixelVsPerceptual}. To demonstrate their differences in optimizing IRNet intuitively, we also visualize the visual attention map produced by the spatial and channel attention layer. As Fig. \ref{FigPixelVsPerceptual} shows, the pixel-wise loss cannot encourage the IRNet to learn a good retargeting model because it regards all pixels in the input image as the same importance. The resulting attention map is sparse and often focuses on the edges of objects. In contrast, the perceptual loss enables the IRNet to focus on integral objects, yielding high-quality target images. Note that the entire training and testing process of our IRNet do not introduce any human annotations or assistance, which is a completely unsupervised learning procedure. Thus, the experiment result well demonstrates the effectiveness of our Circle-IR. In addition, we also demonstrate the results of pixel-wise loss plus perceptual loss. We observe that the pixel-wise loss cannot further improve the performance of perceptual loss. Therefore, in the following experiments, if there is no explicit statement, the used retargeting results are the outputs of the IRNet model optimized by perceptual loss. For convenience, we refer to our approach as Cycle-IR.

\begin{figure}[t]
\centering
\includegraphics[width=8.5 cm]{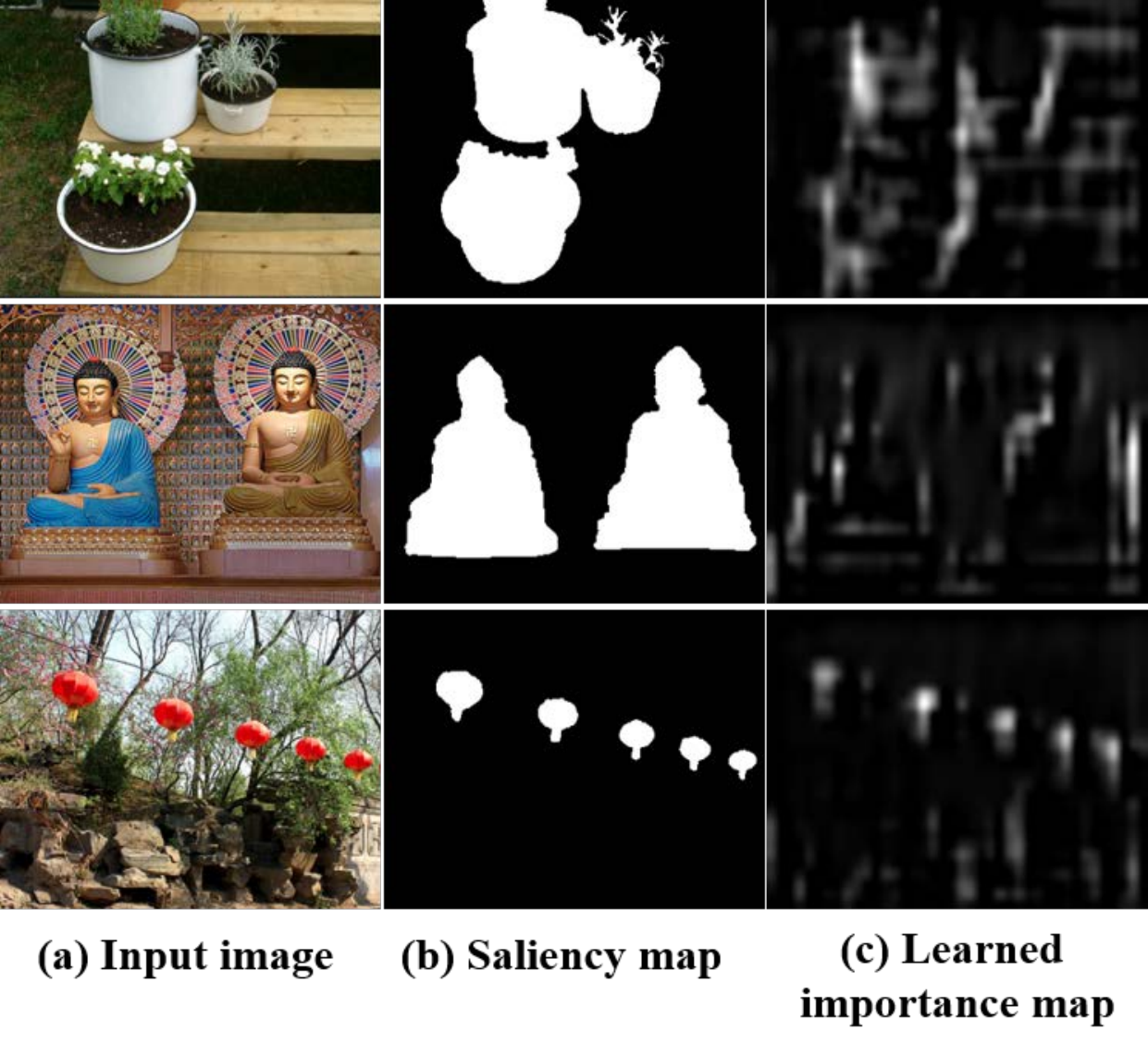}
\caption{Visual comparison between the annotated saliency map and the learned attention map}
\label{FigSaliencyGuidance1}
\end{figure}

\begin{figure}[t]
\centering
\includegraphics[width=8.5 cm]{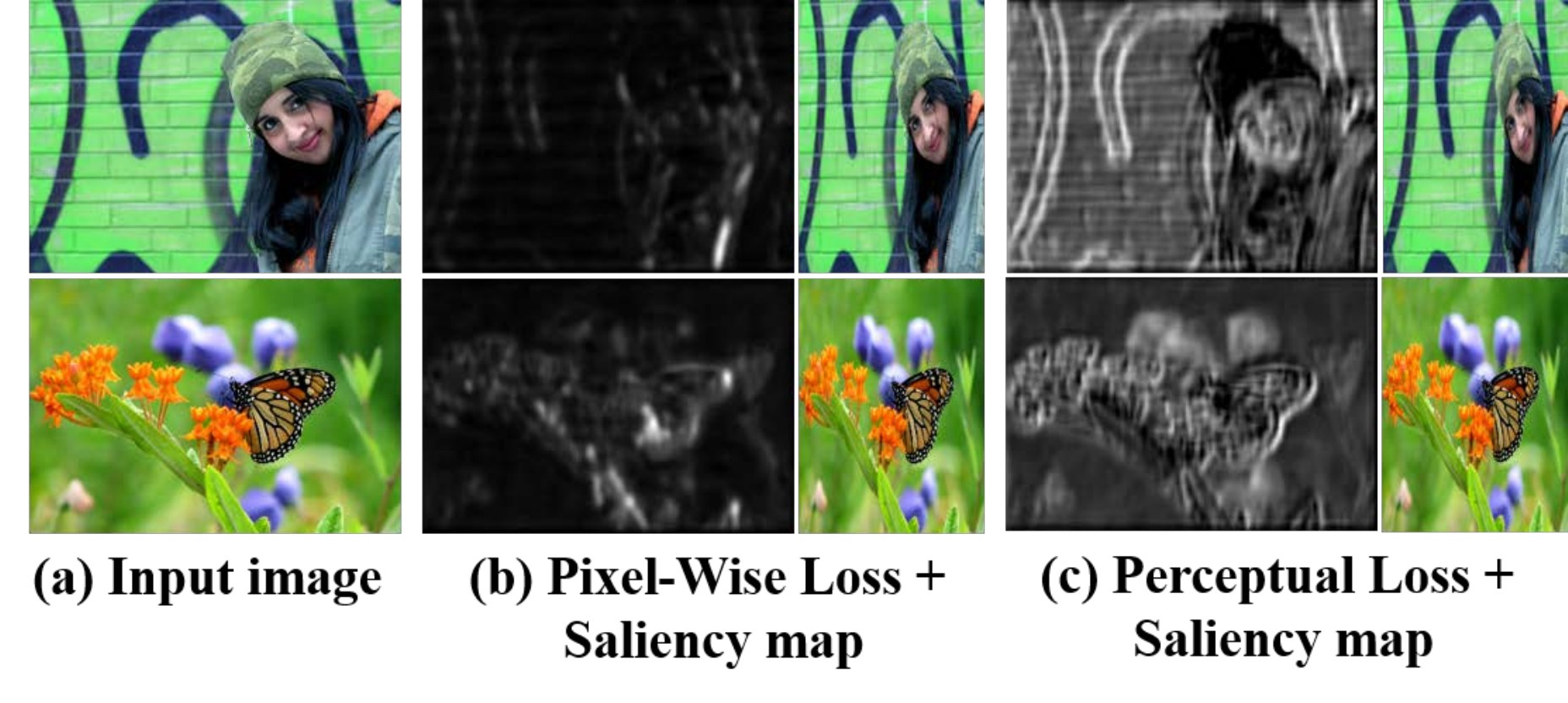}
\caption{Demonstration of integrating saliency maps as guidance with pixel-wise loss or perceptual loss.}
\label{FigSaliencyGuidance2}
\end{figure}

\begin{figure*}[t]
\centering
\includegraphics[width=18 cm]{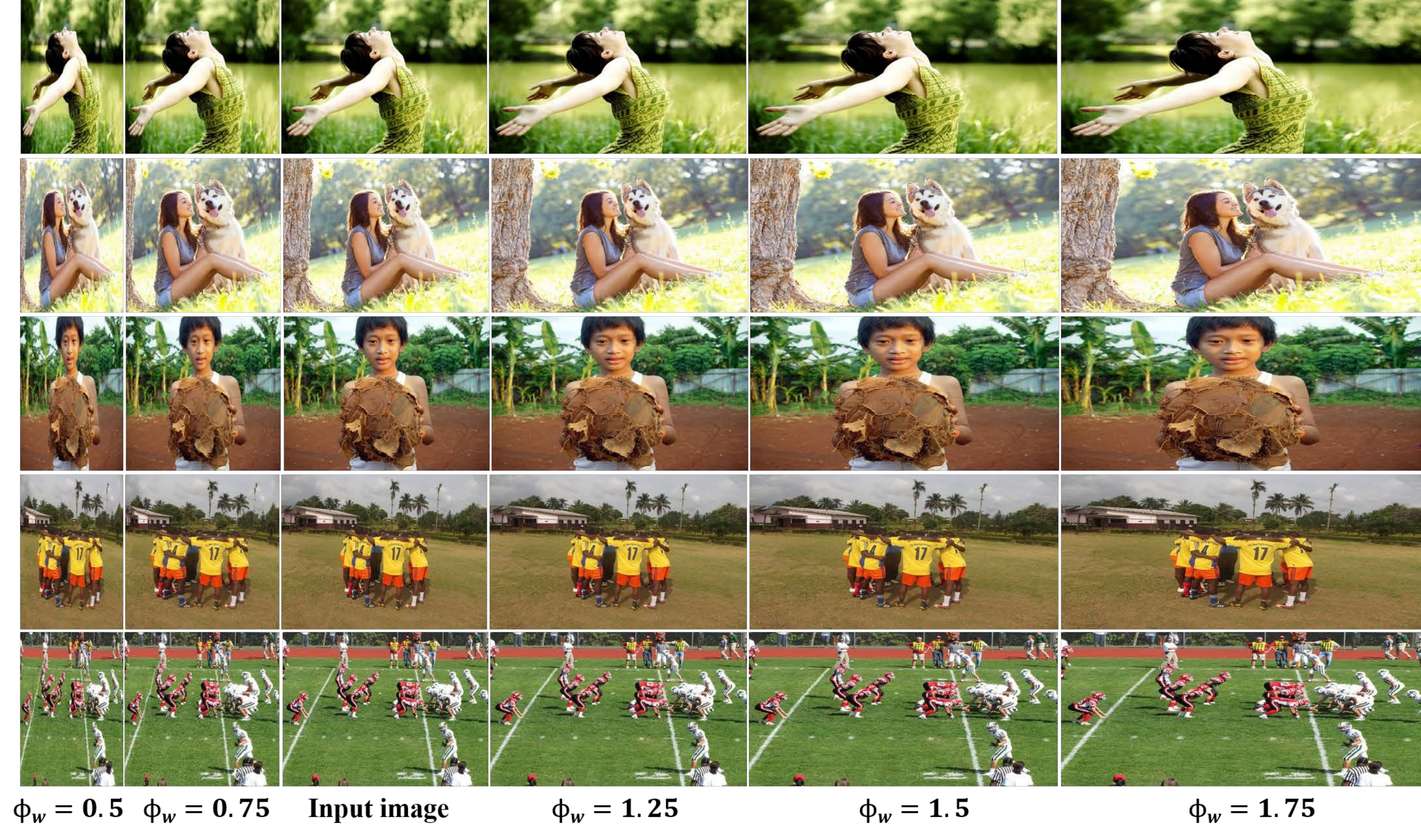}
\caption{Demonstration of the ability of our Cycle-IR to generate target images with arbitrary sizes. Despite the large scale span (from 0.5 to 1.75), our Cycle-IR is able to generate satisfactory target images.}
\label{FigAdjustmentOfAspectRatio}
\end{figure*}

\subsection{Single Cycle Loss vs. Pair Cycle Loss}
\label{SingleCycleVsDoubleCycle}

The proposed cyclic perception coherence loss $\mathcal{L}_{pair}$ (see Eq. (\ref{LossFuntion})) has two terms that correspond to two reverse mappings, respectively. In this section, we experiment with different loss terms to understand the function of each reverse mapping. Here, we use ``Single Cycle Loss (top)'' and ``Single Cycle Loss (bottom)'' to denote the reverse mapping loss term in the top and bottom (see Fig. \ref{FigNetworkArchitecture}), respectively. ``Pair Cycle Loss'' means that both reverse mapping terms are used. Fig. \ref{FigdoubleCycleLoss} illustrates their differences in optimizing IRNet model with several visual examples. Compared with ``Single Cycle Loss (bottom)'' , ``Single Cycle Loss (top)'' plays a more important role in the optimization of IRNet model. This is reasonable because the top reverse mapping focuses on restoring the information removed in the forward retargeting, while the bottom reverse mapping aims at removing the information inserted in the forward retargeting. Therefore, the top branch poses stronger constraints than the bottom one. Furthermore, by integrating both branches, more advanced performance is achieved by ``Pair Cycle Loss''.

\subsection{Saliency Guidance}
\label{SaliencyGuidance}

In this section, we discuss the effect of saliency guidance in optimizing IRNet model. Salient object detection has been studied over the past two decades. Accordingly, a large number of saliency detection datasets are built to promote the development of this field. Here, we carry out an experiment in which the saliency maps provided by HKU-IS dataset are embedded as a priori information in our cycle consistency loss. Figure \ref{FigSaliencyGuidance1} shows the visual comparison between the annotated saliency map and the learned attention map (pixel-wise loss + saliency map). It turns out that with the guidance of annotated saliency map,  the IRNet tends to output sparse attention map, resulting in degrading the retargeting performance. To further understand the effect of the saliency map as guidance, we also incorporate it with cyclic perception loss, as shown in Fig. \ref{FigSaliencyGuidance2}.  We observe that using saliency maps as guidance makes no contribution to both pixel-wise loss and perceptual loss. The possible reason is that the IRNet is not large enough, so it has not learned to detect salient objects in images completely.

\subsection{Zoom In and Out}
\label{AdjustmentOfAspectRatio}

We employ a fully convolutional framework to implement our Cycle-IR approach. Thus, our Cycle-IR is able to deal with input images with arbitrary sizes and generate target images with arbitrary aspect ratios. Figure \ref{FigAdjustmentOfAspectRatio} shows some visual examples of the input aspect ratio $\phi_w$  that equals to 0.5, 0.75, 1.25, 1.5 and 1.75, respectively. Despite the large scale span (from 0.5 to 1.75), our Cycle-IR can well preserve the important areas (such as people, dog, and football) and the overall structure (the white line on the football ground in the fifth row) of the input image.  These visual examples demonstrate the great ability of our Cycle-IR to zoom in and out.

\subsection{Network Size and Computational Efficiency}
\label{NetworkAndEfficiency}

As Fig. \ref{FigNetworkArchitecture} shows, our IRNet model consists of a backbone, three convolutional layers, and a specially designed attention layer. The network parameters of IRNet are about 3.164 M. Such a small amount of network parameters is helpful for fast convergence of the IRNet model. In addition, with our un-optimized TensorFlow implementation on an Nvidia GPU Quadro M4000, our cycle-IR takes 0.365 s for yielding a single image with 1024 $\times$ 768 resolution for resizing to half the original width.

\subsection{Comparison with prior art}
\label{ComparisonWithPriorArt}

In the past decade, numerous representative retargeting approaches have been proposed such as simple scaling operator (SCL), cropping windows (CR), nonhomogeneous warping (WARP \cite{wolf2007nonhomogeneous}), seam carving (SC) \cite{avidan2007seam}, scale-and-stretch (SNS) \cite{wang2008optimized}, Multiop \cite{rubinstein2009multioperator}, shift maps (SM) \cite{pritch2009shiftmap}, streaming video (SV) \cite{krahenbuhl2009a}, and energy-based deformation (LG) \cite{karni2009energybased}. These approaches have published retargeting results on the standard RetargetMe dataset. Thus, we can easily compare with them by evaluating on the RetargetMe dataset. Besides, we also compare with a new retargeting approach \cite{cho2017weakly}, which employs deep network to learn a shift map for manipulating the positions of pixels in the input image. We evaluate the performance of all retargeting approaches in terms of visual quality and user study.

\begin{figure}[t]
\centering
\includegraphics[width=8.5 cm]{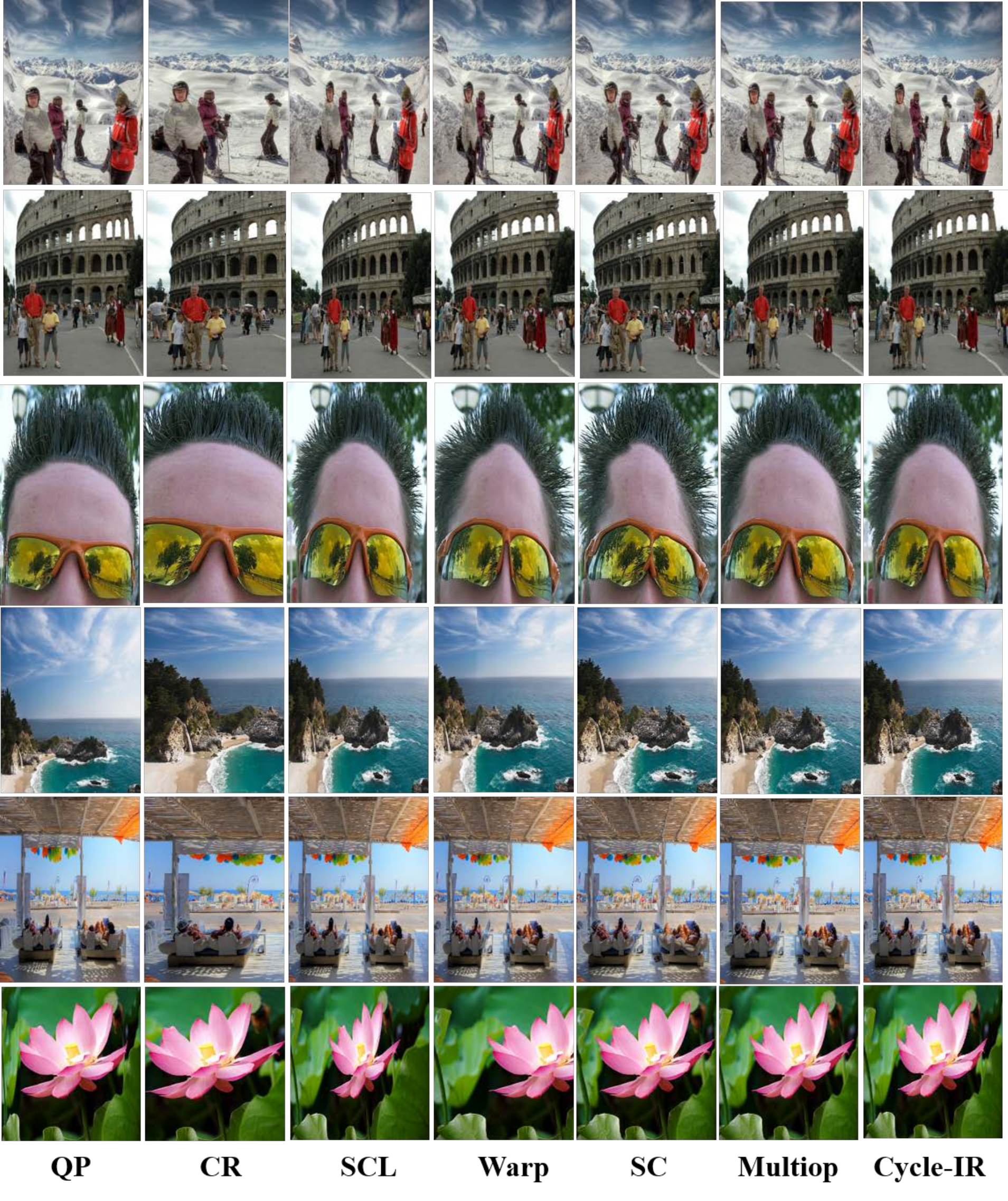}
\caption{Visual comparison of our Cycle-IR with representative retargeting approaches.}
\label{FigComparisonWithPriorArt}
\end{figure}

\subsubsection{Visual Quality}
In this section, we provide a subjective comparison with several representative retargeting approaches and a deep learning based weakly supervised retargeting method. Through the subjective comparison of retargeting results, we will analyze the advantages and disadvantages of each approach in detail.

\textbf{Comparing with Representative Retargeting Approaches}. Fig.\ref{FigComparisonWithPriorArt} shows the visual comparison of our Cycle-IR with several representative retargeting approaches by resizing input images to its half width on RetargetMe \cite{rubinstein2010comparative} dataset. The CR method directly removes the content of input images, resulting in losing important information in the input images. The Multiop \cite{rubinstein2009multioperator} approach achieves good results on most scenes, which benefits from combining the merits of different retargeting operates. The QP method \cite{Chen2010ContentawareIR} can preserve the most important information in the input image, but it introduces distortions and artifacts in the target image. The typical SC method\cite{avidan2007seam} may deform important objects when those seams carved cross over the important objects. The SCL just merges adjacent pixels, which results in over reducing important objects. Benefiting from the powerful ability of deep learning and the effectiveness of cycle consistency, our Cycle-IR is able to produce high-quality target images.

\begin{figure}[t]
\centering
\includegraphics[width=8 cm]{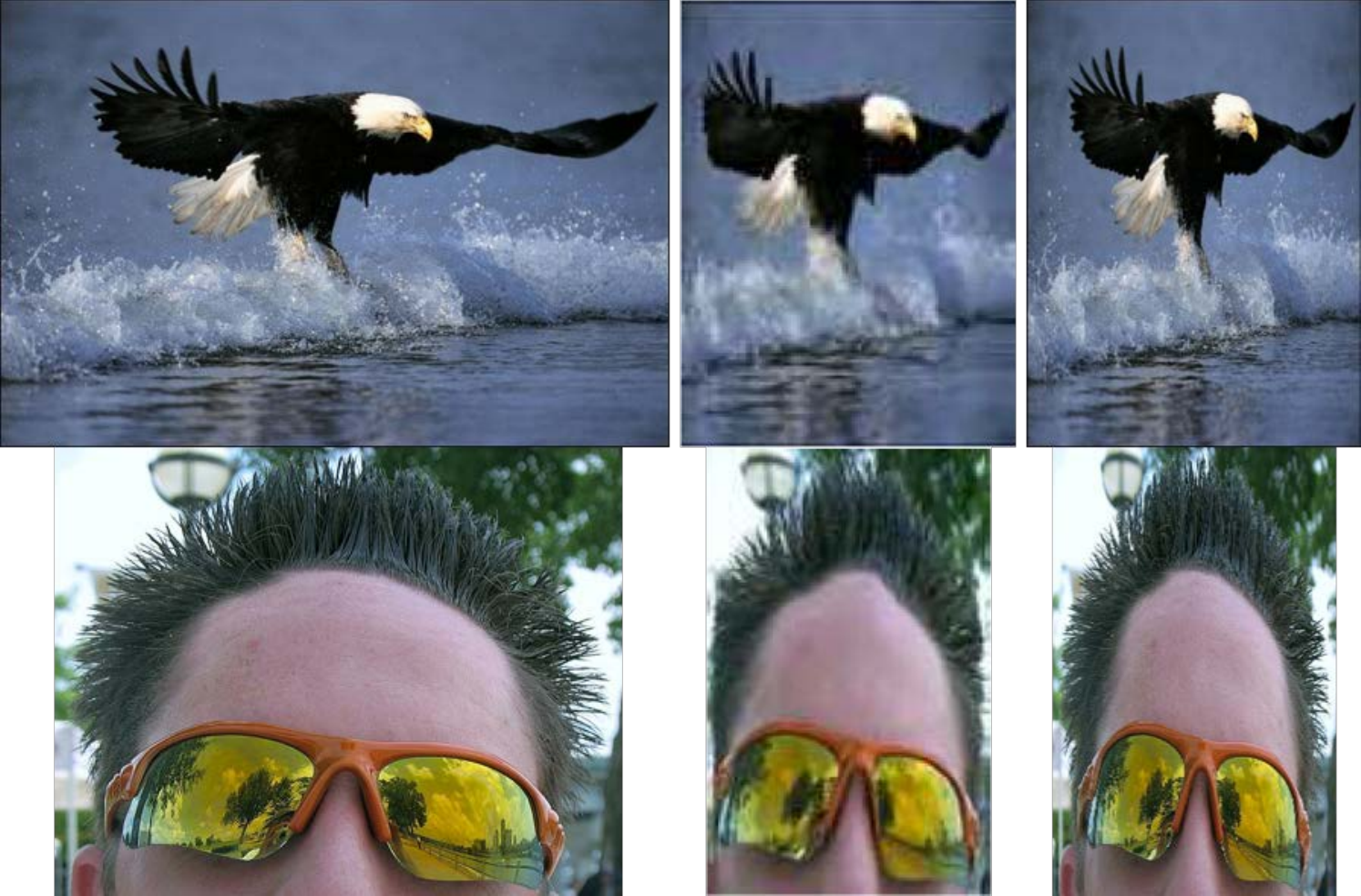}
\caption{From left to right is the original image, Cho $ et~al $. \cite{cho2017weakly} method, our Cycle-IR approach. The original images are retargeted to their half width.}
\label{FigWeaklyDL}
\end{figure}

\textbf{Comparing with Weakly Supervised Deep Retargeting Approach}. Recently, Cho $ et~al $. \cite{cho2017weakly} propose a weakly supervised image retargeting approach. Their deep network is trained on Pascal VOC 2007 dataset \cite{ everingham2010the} with image-level annotations. Fig. \ref{FigWeaklyDL} demonstrates the comparison result of two images, which is quoted from their published paper. From Fig. \ref{FigWeaklyDL} we can observe that comparing with the method \cite{cho2017weakly}, our Cycle-IR can better preserve important objects in the input images. For example, in \cite{cho2017weakly} method, the men's hair in ``glass'' image is reduced too much, while our approach preserves hair better. Besides, the men's forehead has some distortion in \cite{cho2017weakly} method. For ``eagle'' image, our approach preserves better for hawk's wings than\cite{cho2017weakly} method.

\begin{figure*}[t]
\centering
\includegraphics[width=18 cm]{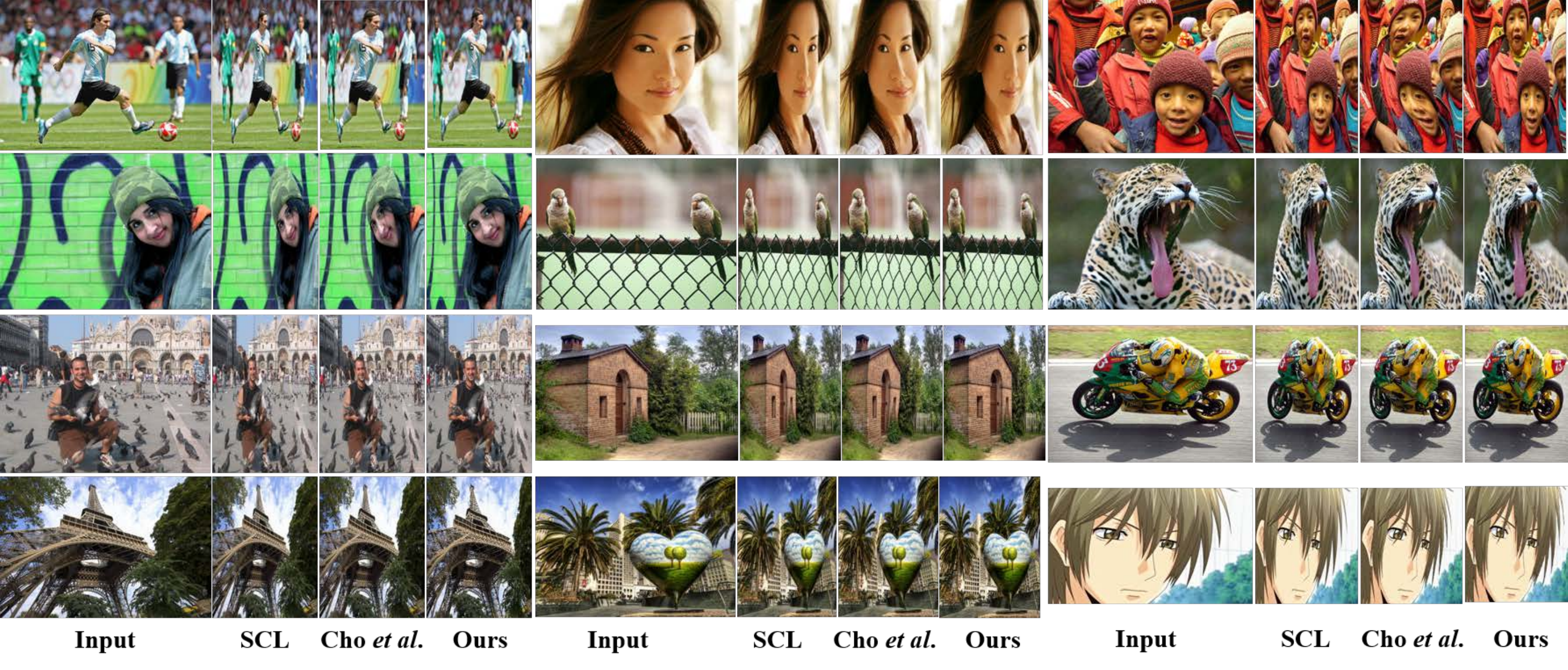}
\caption{We implement Cho $et~al$. \cite{cho2017weakly} method. Thus, more visual comparisons on RetargetMe dataset are demonstrated.}
\label{FigWeaklyDL2}
\end{figure*}

For comprehensively comparing with Cho $et~al$. \cite{cho2017weakly} method, we implement it using TensorFlow platform. Fig. \ref{FigWeaklyDL2} shows more visual comparisons on RetargetMe dataset. It intuitively shows that significant distortions often appear in the retargeted results of Cho $et~al$. \cite{cho2017weakly} method. In contrast, benefiting from the constraint of cyclic perception consistency, our Cycle-IR avoids this serious problem.

\begin{table*}[t]
\centering
\caption{User study for comparing other representative retargeting approaches objectively. There are 14 participants invited to participate in the user study. In a total of 1,820 comparisons, our Cycle-IR has won 490 user preferences.}
\label{TableUserStudy}
\begin{tabular}{c|cccccccccccccc|c|c}
\hline
\textbf{Method}   & \textbf{$P_1$} & \textbf{$P_2$} & \textbf{$P_3$} & \textbf{$P_4$} & \textbf{$P_5$} & \textbf{$P_6$} & \textbf{$P_7$} & \textbf{$P_8$} & \textbf{$P_9$} & \textbf{$P_{10}$} & \textbf{$P_{11}$} & \textbf{$P_{12}$} & \textbf{$P_{13}$} & \textbf{$P_{14}$} & \textbf{Total} & \textbf{Prefer}  \\ \hline
Warp\cite{wolf2007nonhomogeneous}              & 15          & 21          & 24          & 13          & 17          & 16          & 19          & 13          & 17          & 15           & 14           & 19           & 10           & 15           & 228            & 12.52\%          \\
SCL               & 30          & 15          & 17          & 15          & 13          & 23          & 18          & 26          & 27          & 25           & 21           & 18           & 22           & 28           & 298            & 16.37\%          \\
SC \cite{avidan2007seam}                & 24          & 27          & 30          & 30          & 32          & 30          & 25          & 14          & 29          & 10           & 30           & 28           & 16           & 22           & 347            & 19.06\%          \\
Multiop\cite{rubinstein2009multioperator}           & 30          & 32          & 31          & 37          & 33          & 29          & 36          & 36          & 26          & 37           & 34           & 31           & 37           & 28           & 457            & 25.10\%          \\
\textbf{Cycle-IR} & \textbf{31} & \textbf{35} & \textbf{28} & \textbf{35} & \textbf{35} & \textbf{32} & \textbf{32} & \textbf{41} & \textbf{31} & \textbf{43}  & \textbf{31}  & \textbf{34}  & \textbf{45}  & \textbf{37}  & \textbf{490}   & \textbf{26.92\%} \\ \hline
\end{tabular}
\end{table*}

\subsubsection{User Study}

In this part, we conduct a user study to evaluate the preference of retargeted results objectively. Representative retargeting approaches including SC\cite{avidan2007seam}, SCL, Multiop\cite{rubinstein2009multioperator}, and Warp\cite{wolf2007nonhomogeneous} are selected to participate in this user study, because these approaches have robust retargeting performance in most cases. Besides, we invite 14 participants to evaluate the retargeting performance of each approach in terms of visual quality (small distortions and artifacts) and information preservation. We randomly select 13 images from RetargetMe benchmark. All images are retargeted to half width. To improve the efficiency of user study, we developed an evaluating tool using Python. During the evaluation, one original image and two retargeted images produced by 2 out of 5 approaches (including our Cycle-IR) are randomly shown to participants. Afterward, participants choose one retargeted image they prefer.

This user study requires 1,820 comparisons in total. Each image requires 10 comparisons. Each participant needs to evaluate 10 $\times$ 13 = 130 times. Table \ref{TableUserStudy} shows the statistical result of user study. Among the representative retargeting approaches, the Multiop method obtains the most votes and is preferred in 25.10\% in total comparisons, which is consistent with the evaluation results of most methods. Our Cycle-IR obtains 490 votes, which account for 26.92\% in total comparisons. The user preference of our approach is higher than that of all test methods. This objective evaluation results in user study are consistent with the subjective comparisons.

\begin{figure}[t]
\centering
\includegraphics[width=8.3  cm]{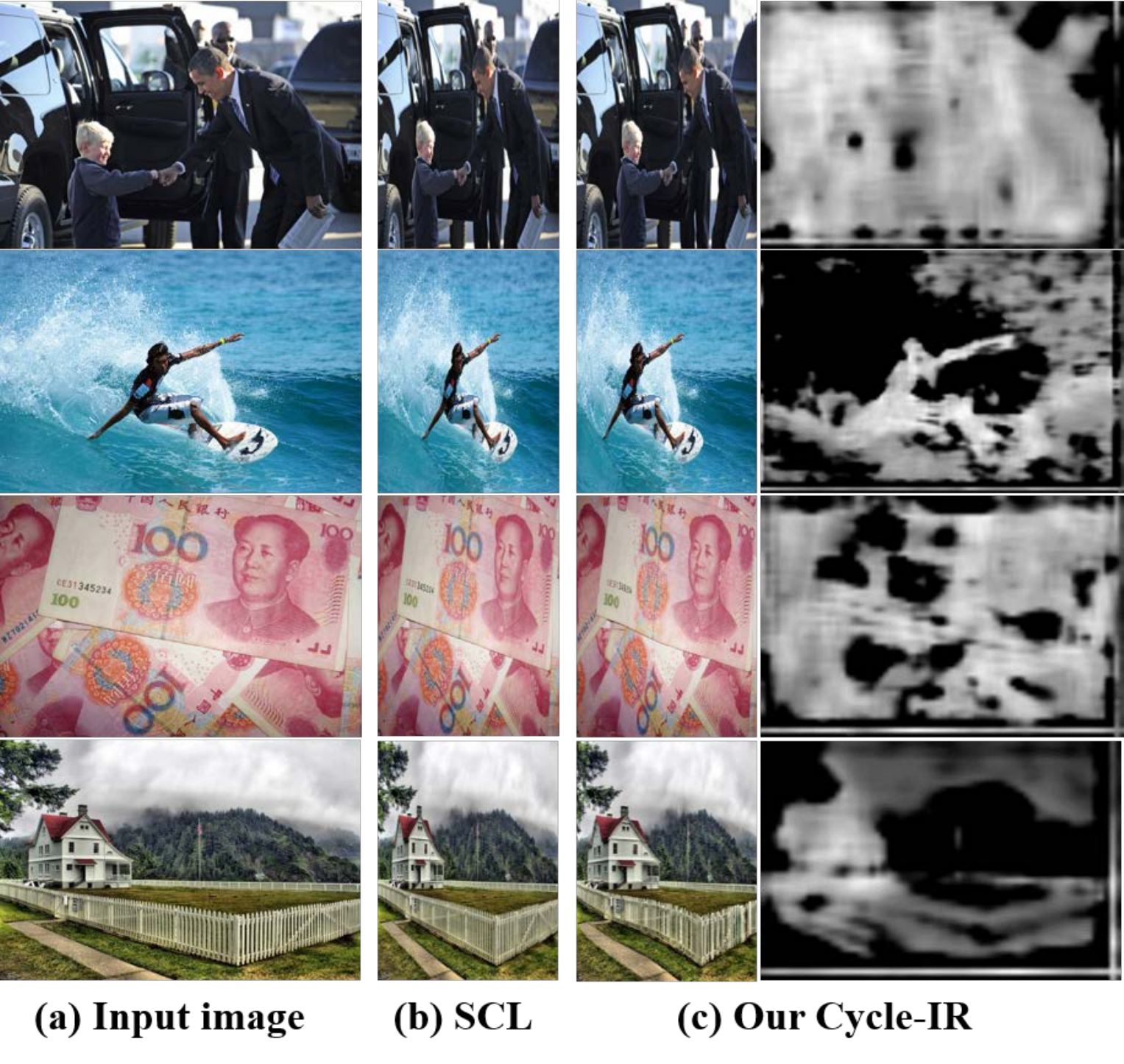}
\caption{Some failure cases.}
\label{FigFailureCases}
\end{figure}

\subsection{Failure Case and Future Work}
\label{FailureCaseAndFutureWork}

Although the proposed Cycle-IR is capable of producing compelling target images in many cases, the results are far from uniformly encouraging. Some failure retargeting results of our approach have been shown in Fig. \ref{FigFailureCases}. In general, these failure cases can be classified into two circumstances. The first one is that the background areas are regarded as the visual importance regions. Typical examples are the images shown in the first and second rows of Fig. \ref{FigFailureCases}. The ``obama'' image is mostly caused by very low contrast, which easily leads to the Cycle-IR failing to distinguish the difference between people (``children'' and ``obama'') and cars. In addition, for the ``surfer'' case, Cycle-IR has a similar retargeting result to SCL method because the sea waves are considered an important area to be preserved. The second type of failure cases is due to the lack of complete attention to important visual areas, as shown in the bottom two rows of Fig. \ref{FigFailureCases}. The important areas of these two images are too large or scattered. Though our Cycle-IR is able to detect some parts of the visually interesting regions, it is still difficult to segment these regions out completely.

There are several possible ways to improve the performance of Cycle-IR. Firstly, a more reasonable loss can make IRNet better optimized to deal with different complex scenes. To better measure the difference of the input images and the reverse-mapped images, we can use a deep network to assess their differences such as using a discriminator. Besides, we can replace the pixel-manipulating step with a learnable component. We think meta-learning is a good choice to get rid of manually designed manipulation. By feeding the desired aspect ratio and the coordinate information of target image, the learnable component outputs a set of parameters such as convolutional kernels. These learned parameters can be used to reconstruct the target images with arbitrary sizes. Finally, deep learning based aesthetic assessment has achieved great success in recent years. Thus, introducing the aesthetic assessment into the optimization of Cycle-IR may help the model to achieve better performance in human visual perception.

\section{Conclusion}

In this paper, we have presented a deep cyclic image retargeting approach that is capable of producing visually pleasing target images given the input images with arbitrary sizes and the desired aspect ratios. We also present a simple yet effective IRNet model to implement the proposed Cycle-IR. Our IRNet model outputs a pair of retargeted images instead of a single one, which is helpful for obtaining a stable retargeting model by exploiting a pair of cycle constraints. To help assist the IRNet in learning accurate attention map, we design a spatial and channel attention layer to discriminate the visual attention of input images effectively. Besides, a cyclic perception coherence loss is introduced to optimize the IRNet model. Extensive experiments including investigation of our Cycle-IR, comparison of other retargeting approaches, and user study demonstrate the superior performance of our approach. More importantly, this work pushes the boundaries of what is possible in the unsupervised setting for content-aware image retargeting.

\bibliographystyle{IEEEtran}


\balance

\end{document}